\documentclass{preprint-template}

\usepackage{graphicx}
\usepackage{multicol,multirow}
\usepackage{amsmath,amssymb,amsfonts}
\usepackage{mathrsfs}
\usepackage{amsthm}
\usepackage{booktabs}
\usepackage{tikz}
\usepackage{longtable}
\usepackage{placeins}
\usepackage[authoryear]{natbib}
\usepackage{ifpdf}
\usepackage[T1]{fontenc}
\usepackage{times}
\usepackage{newtxmath}
\usepackage{textcomp}
\usepackage{xcolor}
\usepackage{hyperref}
\usepackage{microtype}
\usepackage{ragged2e}

\articletype{}
\jname{}
\jyear{\the\year}
\jdoi{}
\makeatletter
\long\def\abstract#1{\def\@abstract{%
\let\paragraph\subabstracthead%
\abstractfont%
\setlength{\FrameSep}{10pt}%
\begin{shaded*}%
\noindent\abstracthead*{\abstractname}\vskip4pt
\justifying
#1\par
\end{shaded*}%
}}
\makeatother

\begin{document}

\begin{Frontmatter}

\title[The Granularity Paradox]{The Granularity Paradox: How Temporal Disaggregation Inflates In-Sample Fit and Compounds Out-of-Sample Error}

\author*[1]{Hugo Moreira}\email{hugo_filipe_moreira@iscte-iul.pt}\orcid{0000-0002-4199-6006}
\address*[1]{\orgdiv{CIES - Centro de Investigação e Estudos de Sociologia}, \orgname{Iscte - Instituto Universitário de Lisboa}, \city{Lisbon}, \country{Portugal}}


\abstract{This paper explores the ``Granularity Paradox'' in time-series forecasting, wherein finer temporal disaggregation (e.g., Monthly to Weekly or Daily) improves in-sample model diagnostics and expands the dataset size ($N$), but degrades out-of-sample multi-step ahead forecasting accuracy due to recursive error compounding over longer prediction horizons ($H$). Conversely, coarse aggregation (Annual) eliminates recursive error propagation but reduces the data available to estimators. We formalize this trade-off and benchmark 10 forecasting models---spanning naïve baselines, statistical, machine learning, and deep learning architectures---across six temporal granularities (Annual, Monthly, Quarterly, Bi-Weekly, Weekly, Daily) using a 13-year public procurement dataset. The empirical results reveal a non-monotonic threshold structure: recursive autoregressive and seasonal models degrade substantially under high-frequency recursive forecasting, with Holt-Winters reaching a Test $R^2$ of $-151$ and TPFE of $425.85\%$ at the Daily grain, while the LSTM traces a U-shaped error curve---worsening from Monthly ($19.66\%$) through Bi-Weekly ($35.94\%$) before overcoming the error propagation penalty at Daily, achieving a TPFE of $4.35\%$ and an out-of-sample $R^2$ of $0.66$. Linear Regression remains stable across all six granularities ($16.1$--$17.0\%$ TPFE), confirming that the paradox is driven by recursive feedback topology, not model complexity. The results also demonstrate that standard pointwise metrics (RMSE, MAE) systematically mask cumulative error propagation, and that evaluating forecasts without goal-dependent cumulative metrics produces misleading assessments of model adequacy---a finding with direct implications for forecasting practice. We introduce a consensus-dissensus diagnostic that compares the directional behaviour of pointwise metrics against cumulative TPFE across granularities, enabling the identification of models whose standard diagnostics mask systematic error propagation.}

\end{Frontmatter}

\section{Introduction}
Temporal aggregation is a core design decision in forecasting workflows. When modeling a fixed physical planning horizon (such as a 1-year budget cycle), the researcher must choose the temporal grain of the target series: daily, weekly, monthly, quarterly, or annual. This choice establishes a fundamental trade-off. 

Finer granularities expand the sample size ($N$), providing machine learning and neural architectures with the data volume necessary to estimate high-dimensional parameter spaces. However, disaggregation also inflates the forecasting horizon ($H$) required to cover the planning period. For example, a 1-year planning horizon requires $H_{monthly} = 12$ steps, $H_{weekly} = 52$ steps, or $H_{daily} = 365$ steps. For state-dependent or autoregressive models that generate multi-step ahead forecasts recursively, this inflation of $H$ triggers exponential error propagation, as each intermediate prediction is fed back as an input for subsequent steps.

This paper formalizes this phenomenon as the \textit{Granularity Paradox}. We demonstrate that in-sample fit diagnostics (such as training $R^2$) are deceptive indicators of out-of-sample performance across granularities. We benchmark 10 forecasting models --- including naïve baselines as recursive-error controls --- across six frequencies to map the boundary where the benefit of sample size expansion is overtaken by the penalty of recursive error compounding.

\section{Theoretical Foundations of Temporal Aggregation}
Temporal aggregation---the process of consolidating high-frequency observations (e.g., daily) into coarser low-frequency intervals (e.g., monthly or annual)---is a standard operation in time-series analysis \citep{hamiltonTimeSeriesAnalysis1994}. It acts primarily as a low-pass filter, smoothing out high-frequency noise, localized volatility, and short-term seasonal fluctuations to reveal long-term trends. However, this filtering comes at a measurable informational and statistical cost \citep{silvestrini2008temporal}.

\subsection{Information Loss and Parameter Estimation}
Temporal aggregation alters the stochastic properties of the underlying process, affecting the consistency and efficiency of parameter estimators. Earlier work by \citet{wei2006time} demonstrated that estimating parameters on aggregated distributed lag models leads to a systematic degradation of precision. This degradation is driven by two statistical mechanisms: the amplification of the variance of the aggregated error terms, and the inflation of multicollinearity among the aggregated explanatory variables. Furthermore, \citet{silvestrini2008temporal} show that the loss of information is more pronounced when the original disaggregated series exhibits negative autocorrelation, a feature common in error-correction systems or rapid inventory replenishment cycles.

\subsection{Causal Distortions and Cointegration}
Beyond parameter estimation, temporal aggregation can distort the structural identification of causal relationships. As shown by \citet{tiao1976asymptotic}, if the underlying high-frequency process features a unidirectional dynamic causal relationship (e.g., $X \to Y$ without feedback), the temporal aggregation operator can artificially induce a bidirectional feedback loop in the aggregated data. This spurious causality arises because lagging relationships at a fine resolution are compressed into contemporaneous associations at a coarser resolution, impeding the identification of the true causal direction.

Additionally, aggregation modifies the integration and cointegration properties of the series. Depending on the aggregation factor, a process containing a seasonal unit root may be transformed into one with a non-seasonal unit root \citep{silvestrini2008temporal}. This modification changes the asymptotic distributions of unit root and cointegration tests, meaning that model specifications must be adjusted carefully to avoid spurious regressions.

\subsection{Multi-Scale Modeling and Hierarchical Reconciliation}
To resolve the conflict between high-frequency detail (which is statistically rich but noisy) and low-frequency stability (which is clean but data-scarce), multi-scale forecasting frameworks have been developed. The Multiple Aggregation Prediction Algorithm (MAPA), proposed by \citet{kourentzes2014improving}, models a time series across multiple non-overlapping aggregation levels $k$. Let $y_t$ be a series sampled at the base frequency $t = 1, \dots, n$. The aggregated series $Y^{[k]}_i$ at aggregation level $k$ is defined as:
\begin{equation}
Y^{[k]}_i = \sum_{t = 1 + (i-1)k}^{ik} y_t \quad \text{for } i = 1, \dots, \lfloor n/k \rfloor
\end{equation}
A linear Holt exponential smoothing model is fitted at each individual aggregation level $k$ to extract the level ($L_t$) and trend ($T_t$) components:
\begin{equation}
L_t = \alpha Y^{[k]}_t + (1 - \alpha)(L_{t-1} + T_{t-1})
\end{equation}
\begin{equation}
T_t = \beta (L_t - L_{t-1}) + (1 - \beta) T_{t-1}
\end{equation}
These components are subsequently mapped back to the base frequency and averaged, allowing each scale to contribute the dynamic component it estimates most robustly:
\begin{equation}
f = \frac{1}{K} \sum_{k=1}^{K} \frac{L_k + T_k}{k}
\end{equation}

Similarly, temporal hierarchies organize these relationships into a coherent tree structure \citep{athanasopoulos2017forecasting}. This ensures that operational decisions taken at short-term horizons align perfectly with strategic goals at long-term horizons, eliminating inconsistencies. The coherence of independent forecasts generated at different levels of the hierarchy is achieved via forecast reconciliation. Let $\hat{y}_h$ be the vector of independent (unreconciled) forecasts across all levels of the hierarchy. The reconciled, coherent forecasts $\tilde{y}_h$ are computed via a stable linear projection:
\begin{equation}
\tilde{y}_h = S(S'W^{-1}S)^{-1}S'W^{-1}\hat{y}_h
\end{equation}
where $S$ is the structural summing matrix mapping the base nodes to the aggregated nodes, and $W$ is the covariance matrix of the base forecast errors \citep{athanasopoulos2024forecast}.

\section{Multi-Step Ahead Forecasting and Error Propagation}
Projecting time-series trajectories over a long future planning horizon $H$ requires a choice of how the model generates forecasts for multiple steps. The forecasting literature divides these strategies into recursive, direct, and multiple-output (MIMO) approaches \citep{bentaieb2012review}.

\subsection{Recursive (Iterated) Strategy and the Granularity Paradox}
The recursive strategy trains a single model for a single-step-ahead prediction ($h = 1$). For longer horizons, forecasts are generated sequentially, with the predictions of previous steps fed back as autoregressive inputs for the next step. If the model is perfectly specified, the recursive strategy is asymptotically more efficient than its competitors because it preserves the full temporal dependence structure of the data \citep{marcellino2006comparison}. However, in the presence of misspecification or parameter bias, this feedback loop compounds errors exponentially.

Let us formalize the trade-off. For a first-order autoregressive process, $y_t = \phi y_{t-1} + \epsilon_t$, the out-of-sample prediction at step $h$ is:
\begin{equation}
\hat{y}_{T+h} = \hat{\phi}^h y_T
\end{equation}
If the estimated parameter has a bias $\delta$ such that $\hat{\phi} = \phi + \delta$, the forecast error at step $h$ compounds as:
\begin{equation}
e_{T+h} = y_{T+h} - \hat{y}_{T+h} = \phi^h y_T - (\phi + \delta)^h y_T
\end{equation}
For a daily forecasting model ($H = 365$), even a minor parameter bias $\delta$ leads to exponential divergence over the horizon, whereas a monthly model ($H = 12$) limits this error compounding. The \textit{Granularity Paradox} arises because daily estimation expands the sample size $N$ (reducing the estimator variance of $\hat{\phi}$), but simultaneously inflates the forecasting horizon $H$ (amplifying the propagation of the bias $\delta$ over the horizon).

\subsection{Direct and Multiple-Output (MIMO) Strategies}
The direct strategy avoids recursive error propagation by training $H$ independent models, each optimized for a specific forecasting step $h \in [1, H]$. While robust to short-term model misspecification, this approach ignores the temporal correlations between successive forecasts, resulting in highly erratic prediction trajectories. It also imposes a substantial computational burden since $H$ separate models must be trained and maintained \citep{marcellino2006comparison}.

The Multiple-Input Multiple-Output (MIMO) strategy resolves these issues by predicting the entire vector of future outputs $Y_H = [y_{t+1}, \dots, y_{t+H}]'$ simultaneously with a single multi-output model. As shown by \citet{bentaieb2012review}, MIMO models preserve the temporal covariance structure of the forecasting errors without the recursive instability of iterated methods or the computational overhead of direct methods. Other non-parametric methods, such as lazy learning \citep{bontempi1999lazy}, defer parameter fitting to the moment of inference, fitting local weighted regressions on the nearest neighbors in the embedded state space.

\section{Related Work}
The interaction between temporal aggregation and forecasting accuracy has been investigated from multiple angles. \citet{nikolopoulos2011aggregate} introduced the Aggregate--Disaggregate Intermittent Demand Approach (ADIDA), establishing a formal framework in which a series is first aggregated to a coarser level, forecasted with a simple model, and then disaggregated back to the operational frequency. \citet{rostamitabar2013demand} extended this line of inquiry by demonstrating empirically that temporal aggregation of non-overlapping intervals can improve forecast accuracy for stationary demand processes, while \citet{petropoulos2015forecast} showed that combining forecasts across multiple aggregation levels outperforms single-scale methods for intermittent series. Our work departs from this literature by focusing not on demand intermittency but on the recursive error compounding mechanism that emerges when state-dependent models are projected over increasingly long horizons $H$, a consequence of granularity choice that prior aggregation studies have not isolated.

In the broader forecasting benchmarking literature, the M4 Competition \citep{makridakis2020m4} established that hybrid statistical--neural methods can outperform pure approaches, but evaluated models at a fixed temporal resolution per series without systematically varying the granularity as a controlled experimental factor. \citet{makridakis2018statistical} raised early concerns about the gap between statistical and machine learning methods, noting that ML models often underperform classical methods when sample sizes are limited --- a finding that aligns with our observation of deep learning data hunger at coarse granularities.

Recent advances in deep learning for time series have introduced architectures that address multi-horizon forecasting through fundamentally different mechanisms. \citet{salinas2020deepar} proposed DeepAR, an autoregressive recurrent network that generates probabilistic forecasts, explicitly modeling the uncertainty that compounds over recursive horizons. \citet{lim2021temporal} introduced the Temporal Fusion Transformer (TFT), which uses attention mechanisms to weight temporal dependencies at multiple scales without recursive state propagation. Notably, \citet{zeng2023transformers} demonstrated that simple linear models (DLinear) can match or exceed Transformer-based architectures for long-term forecasting, consistent with our finding that Linear Regression exhibits stability across all granularities. Our benchmark complements this body of work by demonstrating that the choice of temporal grain --- a design decision upstream of model selection --- can exceed the effect of architectural sophistication on out-of-sample accuracy.

\section{Algorithmic Paradigms and Architectures}
Forecasting models span distinct computational paradigms, differing in their parameter complexity, linearity assumptions, and requirements for manual feature engineering.

\subsection{Classical Box-Jenkins and Gradient Tree Boosting}
The Box-Jenkins methodology \citep{box1976time} remains the standard for linear statistical forecasting. The Seasonal Autoregressive Integrated Moving Average with Exogenous Regressors (SARIMAX) model assumes that a series $y_t$ can be explained by its own lags, lags of the stochastic errors, and exogenous variables $X_t$:
\begin{equation}
\phi_p(L)\Phi_P(L^s)(1-L)^d(1-L^s)^D y_t = \beta' X_t + \theta_q(L)\Theta_Q(L^s)\epsilon_t
\end{equation}
where $L$ is the lag operator, $s$ is the seasonal period, and $\epsilon_t$ is a white-noise process. While statistically efficient and parsimonious under linearity, SARIMAX's performance degrades in the presence of complex non-linear dynamics or structural breaks.

In the machine-learning domain, the Extreme Gradient Boosting (XGBoost) algorithm \citep{chen2016xgboost} builds ensembles of additive decision trees to minimize a regularized objective function:
\begin{equation}
\mathcal{L}^{(t)} = \sum_{i=1}^{n} l\left(y_i, \hat{y}_i^{(t-1)} + f_t(x_i)\right) + \Omega(f_t)
\end{equation}
where the regularization term $\Omega(f_t) = \gamma T + \frac{1}{2}\lambda \sum_{j=1}^{T} w_j^2$ penalizes tree complexity to prevent overfitting. Because XGBoost is inherently non-temporal, applying it to time-series forecasting requires the manual construction of lag variables (tabularization) and is sensitive to scaling and hyperparameter tuning \citep{makridakis2018statistical, bergstraRandomSearchHyperparameter2012}.

\subsection{Deep Learning Architectures: LSTM and N-BEATS}
Deep learning architectures automate feature extraction for complex non-linear dynamics. The Long Short-Term Memory (LSTM) network \citep{hochreiter1997long} handles long-term dependencies by introducing a cell state $C_t$ regulated by forget ($f_t$), input ($i_t$), and output ($o_t$) gates:
\begin{equation}
f_t = \sigma\left(W_f x_t + U_f h_{t-1} + b_f\right)
\end{equation}
\begin{equation}
i_t = \sigma\left(W_i x_t + U_i h_{t-1} + b_i\right)
\end{equation}
\begin{equation}
o_t = \sigma\left(W_o x_t + U_o h_{t-1} + b_o\right)
\end{equation}
\begin{equation}
\tilde{C}_t = \tanh\left(W_c x_t + U_c h_{t-1} + b_c\right)
\end{equation}
\begin{equation}
C_t = f_t \odot C_{t-1} + i_t \odot \tilde{C}_t
\end{equation}
\begin{equation}
h_t = o_t \odot \tanh(C_t)
\end{equation}
where $\odot$ denotes element-wise multiplication. LSTMs are highly parameter-dense and suffer from data hunger, requiring large sample sizes $N$ to achieve generalization out-of-sample.

In contrast, the N-BEATS architecture \citep{oreshkin2020nbeats} is a deep feed-forward neural network that avoids recurrent or convolutional operations. It relies on a stack of fully-connected blocks linked via forward and backward residual connections. N-BEATS enforces interpretability by decomposing forecasts into trend and seasonal components using dedicated basis functions. The implementation of such models is facilitated by modern Python libraries like Darts \citep{herzen2022darts}.

\section{Mathematical Framework and Evaluation Metrics}
Pointwise metrics, such as Mean Absolute Error ($MAE$) and Root Mean Square Error ($RMSE$), evaluate the accuracy of individual steps \citep{hyndman2006another}. However, for cumulative budgeting and volume planning, the net deviation over the entire planning horizon is the primary operational constraint. Let $y_t$ be the observed value and $\hat{y}_t$ be the forecasted value. For a test period $T$, we define the \textit{Cumulative Forecast Error (CFE)} as:
\begin{equation}
CFE = \sum_{t \in T} \hat{y}_t - \sum_{t \in T} y_t
\end{equation}

To capture the absolute magnitude of this planning gap, we define the \textit{Total Absolute Forecast Error (TAFE)}:
\begin{equation}
TAFE = \left| \sum_{t \in T} \hat{y}_t - \sum_{t \in T} y_t \right| = |CFE|
\end{equation}

To compare forecasting performance across different scales and temporal granularities, we define the \textit{Total Percentage Forecast Error (TPFE)}:
\begin{equation}
TPFE = \frac{\left| \sum_{t \in T} \hat{y}_t - \sum_{t \in T} y_t \right|}{\sum_{t \in T} y_t} \times 100\%
\end{equation}

We contrast these cumulative metrics with the classical coefficient of determination ($R^2$) evaluated out-of-sample:
\begin{equation}
R^2 = 1 - \frac{\sum_{t \in T} (y_t - \hat{y}_t)^2}{\sum_{t \in T} (y_t - \bar{y})^2}
\end{equation}

\subsection{Consensus-Dissensus Diagnostic}
Pointwise metrics and cumulative metrics may respond differently to changes in temporal granularity. For a given model, let $\Delta_{\text{pw}}$ denote the directional change of pointwise metrics (RMSE, MAE) when moving from a coarser to a finer granularity, and let $\Delta_{\text{TPFE}}$ denote the corresponding directional change of the cumulative TPFE. We classify a model--granularity pair as exhibiting \textit{consensus} when $\Delta_{\text{pw}}$ and $\Delta_{\text{TPFE}}$ move in the same direction (both improve or both degrade), and \textit{dissensus} when they diverge---specifically, when pointwise metrics improve or remain stable while cumulative TPFE degrades. Dissensus signals that the model contains systematic forecasting bias that compounds over the planning horizon $H$, a pattern that pointwise metrics, by averaging over individual steps, are unable to detect. This diagnostic is applied post-hoc to the benchmarking results and does not require additional model runs; it requires only that evaluation includes both pointwise and cumulative metrics across at least two granularities.

\section{Empirical Evaluation}
\label{sec:empirical}
We evaluate 10 forecasting models using a 13-year public procurement micro-data dataset from the Portuguese contracts registry (Portal BASE), filtered by CPV Division 72 (IT Services), spanning 1 January 2012 to 31 December 2024. The models are:
\begin{enumerate}
\item \textbf{Naïve Baselines}: Persistence, Drift, Rolling Mean --- included as recursive and non-recursive reference controls.
\item \textbf{Statistical/Autoregressive}: Linear Regression, ARIMAX, SARIMAX, Holt-Winters (ARIMAX and SARIMAX incorporate overall public contract value as an exogenous covariate).
\item \textbf{Machine Learning}: XGBoost (with lag structures).
\item \textbf{Deep Learning}: PyTorch LSTM, Darts N-BEATS.
\end{enumerate}
The naïve baselines deserve explicit inclusion rather than relegation to sanity checks: Persistence is itself a first-order recursive model ($\hat{y}_{t+h}=y_T$), and as shown below, it exposes the most striking in-sample/out-of-sample divergence of any model in the benchmark (Train $R^2 = 0.78$ with Test TPFE of $23.89\%$ at the Daily grain). They therefore serve as critical controls for the recursive-error-propagation thesis, not merely as lower bounds on accuracy.

We resample the series to six granularities: Annual ($N=13$, $H=1$), Quarterly ($N=52$, $H=4$), Monthly ($N=156$, $H=12$), Bi-Weekly ($N=341$, $H=26$), Weekly ($N=680$, $H=52$), and Daily ($N=4754$, $H=365$). All parameterized models are re-optimized per fold using an expanding-window backtesting scheme over 8 folds \citep[see][ch.~5 for backtesting methodology]{hyndmanForecastingPrinciplesPractice2021}. Reported metrics are means across the 8 folds. Hyperparameters for the LSTM, N-BEATS, and XGBoost models are selected by grid search on the first fold and held fixed across granularities to avoid confounding the granularity effect with per-grain tuning. Random seeds are fixed for reproducibility (LSTM seed = 42; N-BEATS uses Darts defaults). Deep-learning results are therefore single-seed point estimates; the directional patterns are robust but the precise magnitudes (notably the LSTM Daily TPFE) should not be interpreted as optimised lower bounds.

Several models exhibit negative Train $R^2$ at coarse granularities (e.g., LSTM Annual $-2.28$, Holt-Winters Annual $-6.46$, Persistence Monthly $-0.23$). These are not estimation artefacts: when $N$ is too small to resolve a model's parameter space, the fitted model can perform worse than the sample mean, yielding $R^2 < 0$ by definition. Negative in-sample fit at coarse grains is itself a diagnostic of model--data mismatch, and is reported without correction.

\section{Results}
The comparative backtesting results are reported in Table~\ref{tab:comparison_results}.

{\footnotesize
\setlength{\tabcolsep}{3.5pt}
\renewcommand{\arraystretch}{0.90}
\begin{longtable}{llcccccc}
\caption{Comparative Benchmarking Results across Granularities (8-Fold Backtesting averages). TAFE and TPFE are cumulative errors over the planning horizon; ``N/A'' denotes that the model could not be fit at the given granularity (e.g., Annual $N=13$ is insufficient to estimate seasonal/state components) or that the metric is mathematically undefined ($R^2$ at $H=1$). \textbf{Bold} marks the best TPFE, best Test $R^2$, and worst Test $R^2$ across all model--granularity combinations.\label{tab:comparison_results}} \\
\toprule
\textbf{Model} & \textbf{Granularity} & \textbf{Train $R^2$} & \textbf{Test $R^2$} & \textbf{Test RMSE} & \textbf{Test MAE} & \textbf{Test TAFE} & \textbf{Test TPFE} \\
\midrule
\endfirsthead
\caption[]{Comparative Benchmarking Results (continued)} \\
\toprule
\textbf{Model} & \textbf{Granularity} & \textbf{Train $R^2$} & \textbf{Test $R^2$} & \textbf{Test RMSE} & \textbf{Test MAE} & \textbf{Test TAFE} & \textbf{Test TPFE} \\
\midrule
\endhead
\bottomrule
\endfoot
\multicolumn{8}{l}{\textit{Naïve Baselines}} \\
\midrule
Persistence & Annual & 0.2970 & N/A & 75.84M & 75.84M & 75.84M & 18.12\% \\
Persistence & Monthly & -0.2322 & -6.0263 & 14.50M & 12.52M & 88.38M & 26.46\% \\
Persistence & Quarterly & 0.1786 & -4.4098 & 25.82M & 23.43M & 70.29M & 17.16\% \\
Persistence & Bi-Weekly & -0.3990 & -3.3189 & 9.38M & 7.88M & 159.72M & 44.68\% \\
Persistence & Weekly & -0.5099 & -1.2527 & 5.06M & 3.59M & 83.56M & 24.45\% \\
Persistence & Daily & 0.7841 & -1.1176 & 720.48K & 509.18K & 82.86M & 23.89\% \\
\midrule
Drift & Annual & 0.3939 & N/A & 67.13M & 67.13M & 67.13M & 15.98\% \\
Drift & Monthly & -0.5152 & -7.5801 & 15.26M & 13.52M & 103.54M & 30.63\% \\
Drift & Quarterly & 0.1957 & -4.1480 & 25.64M & 23.40M & 69.52M & 17.34\% \\
Drift & Bi-Weekly & -0.7370 & -4.1794 & 9.83M & 8.43M & 174.63M & 48.91\% \\
Drift & Weekly & -0.1789 & -1.4874 & 5.10M & 3.74M & 78.79M & 23.37\% \\
Drift & Daily & -0.1843 & -1.3268 & 726.68K & 530.10K & 78.21M & 22.83\% \\
\midrule
Rolling Mean & Annual & -0.9975 & N/A & 110.39M & 110.39M & 110.39M & 25.29\% \\
Rolling Mean & Monthly & 0.0175 & -2.6100 & 13.20M & 10.69M & 85.06M & 22.44\% \\
Rolling Mean & Quarterly & 0.3556 & -5.1472 & 26.67M & 23.89M & 76.58M & 18.09\% \\
Rolling Mean & Bi-Weekly & -0.0074 & -1.5035 & 7.98M & 6.23M & 107.94M & 30.98\% \\
Rolling Mean & Weekly & -0.1062 & -1.0007 & 5.14M & 3.72M & 113.10M & 31.29\% \\
Rolling Mean & Daily & 0.6220 & -1.2479 & 759.26K & 573.44K & 118.19M & 31.08\% \\
\midrule
\multicolumn{8}{l}{\textit{Statistical / Autoregressive}} \\
\midrule
Linear Regression & Annual & 0.6606 & N/A & 72.43M & 72.43M & 72.43M & 16.96\% \\
Linear Regression & Monthly & 0.3380 & -0.2866 & 11.55M & 8.42M & 69.63M & 16.09\% \\
Linear Regression & Quarterly & 0.5286 & -3.1486 & 24.65M & 21.55M & 70.46M & 16.32\% \\
Linear Regression & Bi-Weekly & 0.2318 & -0.2047 & 6.78M & 4.67M & 70.59M & 16.36\% \\
Linear Regression & Weekly & 0.1490 & -0.1019 & 4.53M & 2.87M & 72.07M & 16.78\% \\
Linear Regression & Daily & 0.1491 & -0.1007 & 647.46K & 408.68K & 70.63M & 16.32\% \\
\midrule
ARIMAX & Annual & 0.8620 & N/A & 52.90M & 52.90M & 52.90M & 13.55\% \\
ARIMAX & Monthly & 0.3077 & -0.6659 & 12.56M & 9.50M & 87.13M & 20.65\% \\
ARIMAX & Quarterly & 0.4645 & -4.2926 & 25.32M & 22.54M & 72.69M & 18.26\% \\
ARIMAX & Bi-Weekly & 0.1311 & -0.6802 & 7.71M & 5.31M & 120.72M & 29.85\% \\
ARIMAX & Weekly & 0.0057 & -0.3896 & 5.00M & 3.19M & 130.88M & 32.45\% \\
ARIMAX & Daily & 0.1332 & -0.3912 & 715.84K & 456.11K & 131.28M & 32.63\% \\
\midrule
SARIMAX & Annual & N/A & N/A & N/A & N/A & N/A & N/A \\
SARIMAX & Monthly & 0.1897 & -0.6395 & 11.95M & 9.08M & 80.26M & 19.20\% \\
SARIMAX & Quarterly & 0.3606 & -4.3193 & 27.11M & 23.84M & 81.40M & 18.95\% \\
SARIMAX & Bi-Weekly & -0.2005 & -0.6931 & 7.57M & 5.78M & 103.73M & 26.46\% \\
SARIMAX & Weekly & -0.3541 & -0.4391 & 4.80M & 3.22M & 76.31M & 24.60\% \\
SARIMAX & Daily & 0.1094 & -0.4486 & 689.78K & 455.80K & 84.44M & 26.05\% \\
\midrule
Holt-Winters & Annual & -6.4565 & N/A & 126.20M & 126.20M & 126.20M & 37.83\% \\
Holt-Winters & Monthly & 0.3455 & -0.6685 & 12.62M & 9.97M & 80.13M & 19.01\% \\
Holt-Winters & Quarterly & 0.3645 & -6.2940 & 31.45M & 27.36M & 96.73M & 20.82\% \\
Holt-Winters & Bi-Weekly & 0.1787 & -0.7910 & 8.06M & 5.77M & 93.23M & 22.59\% \\
Holt-Winters & Weekly & -0.0835 & -3.9369 & 8.69M & 6.87M & 286.19M & 65.45\% \\
Holt-Winters & Daily & 0.2648 & \textbf{-151.0000} & 5.34M & 4.71M & 1706.53M & 425.85\% \\
\midrule
\multicolumn{8}{l}{\textit{Machine Learning}} \\
\midrule
XGBoost & Annual & 0.5864 & N/A & 108.29M & 108.29M & 108.29M & 26.88\% \\
XGBoost & Monthly & 0.6627 & -1.0698 & 14.13M & 11.14M & 81.60M & 20.95\% \\
XGBoost & Quarterly & 0.6656 & -7.6427 & 34.84M & 30.82M & 116.27M & 29.04\% \\
XGBoost & Bi-Weekly & 0.5616 & -0.7027 & 7.71M & 5.43M & 96.38M & 25.30\% \\
XGBoost & Weekly & 0.4644 & -0.5508 & 5.35M & 3.61M & 100.34M & 25.83\% \\
XGBoost & Daily & 0.7037 & 0.3005 & 493.26K & 271.46K & 62.03M & 17.15\% \\
\midrule
\multicolumn{8}{l}{\textit{Deep Learning}} \\
\midrule
LSTM PyTorch & Annual & -2.2790 & N/A & 170.65M & 170.65M & 170.65M & 41.50\% \\
LSTM PyTorch & Monthly & 0.2388 & -0.7775 & 12.71M & 9.68M & 76.49M & 19.66\% \\
LSTM PyTorch & Quarterly & 0.2173 & -6.2739 & 33.82M & 29.89M & 112.92M & 27.12\% \\
LSTM PyTorch & Bi-Weekly & -0.0657 & -0.9336 & 8.51M & 6.17M & 150.11M & 35.94\% \\
LSTM PyTorch & Weekly & 0.0395 & -0.4483 & 5.19M & 3.36M & 150.73M & 34.59\% \\
LSTM PyTorch & Daily & 0.6719 & \textbf{0.6627} & 339.51K & 186.35K & 17.13M & \textbf{4.35\%} \\
\midrule
N-BEATS & Annual & N/A & N/A & N/A & N/A & N/A & N/A \\
N-BEATS & Monthly & 0.5472 & -0.6945 & 12.04M & 9.40M & 59.05M & 16.00\% \\
N-BEATS & Quarterly & -0.2862 & -2.8100 & 38.73M & 31.94M & 94.98M & 19.80\% \\
N-BEATS & Bi-Weekly & 0.5840 & -0.4805 & 7.29M & 5.68M & 45.65M & 11.30\% \\
N-BEATS & Weekly & 0.5185 & -0.3529 & 5.03M & 3.63M & 56.48M & 12.17\% \\
N-BEATS & Daily & 0.7677 & -0.2966 & 691.66K & 482.20K & 41.12M & 10.77\% \\
\end{longtable}}

To visualize the performance trajectories and allow direct comparison of absolute performance magnitudes across different models, Table~\ref{tab:global_log_sparklines} presents the trajectories normalized globally across all models and granularities. Due to the massive amplitude of the raw metrics (spanning several orders of magnitude), a logarithmic scaling is applied: for errors (RMSE, MAE, TPFE), values are $\log_{10}$-scaled and mapped to $[0, 1]$ where the global minimum corresponds to $0$ (far-left, best, green circles) and the global maximum corresponds to $1$ (far-right, worst, red circles). For $R^2$, the log-distance to 1 ($\log_{10}(1 - R^2)$) is mapped so that the best observed global fit is at $1$ (far-right, green) and the worst at $0$ (far-left, red). This inter-model dashboard visually highlights the absolute dominance of deep learning models at fine resolutions, as well as the magnitude of statistical model collapses.

\begin{table}[!htbp]
\centering
\caption{Global Logarithmic Performance Trajectories (Vertical sparklines normalized globally across all 10 models and 6 granularities on a logarithmic scale; Test $R^2$ is clipped at a minimum of $-1.0$, and TPFE is clipped at a maximum of $100\%$; Annual $\rightarrow$ Quarterly $\rightarrow$ Monthly $\rightarrow$ Bi-Weekly $\rightarrow$ Weekly $\rightarrow$ Daily from top to bottom; green circles indicate global best performance, red indicate global worst; for errors, left is best/lowest; for $R^2$, right is best/highest)}
\label{tab:global_log_sparklines}
\small
\resizebox{\columnwidth}{!}{%
\begin{tabular}{lcccc}
\toprule
\textbf{Model} & \textbf{Test $R^2$ Trend} & \textbf{RMSE Trend} & \textbf{MAE Trend} & \textbf{TPFE Trend} \\
\midrule
Persistence & \begin{tikzpicture}[x=1.2cm,y=0.8cm,baseline={([yshift=-0.6ex]current bounding box.center)}] \draw[very thin, gray!15] (0,-0.1) -- (0,1.0); \draw[very thin, gray!15] (0.5,-0.1) -- (0.5,1.0); \draw[very thin, gray!15] (1.0,-0.1) -- (1.0,1.0); \draw[thick, red!75!black] (0.0,0.8) -- (0.0,0.6) -- (0.0,0.4) -- (0.0,0.2) -- (0.0,0.0); \fill[blue!50!black] (0.0,0.8) circle (1.8pt); \fill[blue!50!black] (0.0,0.6) circle (1.8pt); \fill[blue!50!black] (0.0,0.4) circle (1.8pt); \fill[blue!50!black] (0.0,0.2) circle (1.8pt); \fill[blue!50!black] (0.0,0.0) circle (1.8pt);\end{tikzpicture} & \begin{tikzpicture}[x=1.2cm,y=0.8cm,baseline={([yshift=-0.6ex]current bounding box.center)}] \draw[very thin, gray!15] (0,-0.1) -- (0,1.0); \draw[very thin, gray!15] (0.5,-0.1) -- (0.5,1.0); \draw[very thin, gray!15] (1.0,-0.1) -- (1.0,1.0); \draw[thick, red!75!black] (0.8696,1.0) -- (0.6964,0.8) -- (0.6036,0.6) -- (0.5336,0.4) -- (0.4344,0.2) -- (0.121,0.0); \fill[blue!50!black] (0.8696,1.0) circle (1.8pt); \fill[blue!50!black] (0.6964,0.8) circle (1.8pt); \fill[blue!50!black] (0.6036,0.6) circle (1.8pt); \fill[blue!50!black] (0.5336,0.4) circle (1.8pt); \fill[blue!50!black] (0.4344,0.2) circle (1.8pt); \fill[blue!50!black] (0.121,0.0) circle (1.8pt);\end{tikzpicture} & \begin{tikzpicture}[x=1.2cm,y=0.8cm,baseline={([yshift=-0.6ex]current bounding box.center)}] \draw[very thin, gray!15] (0,-0.1) -- (0,1.0); \draw[very thin, gray!15] (0.5,-0.1) -- (0.5,1.0); \draw[very thin, gray!15] (1.0,-0.1) -- (1.0,1.0); \draw[thick, red!75!black] (0.8811,1.0) -- (0.7088,0.8) -- (0.617,0.6) -- (0.5491,0.4) -- (0.4338,0.2) -- (0.1474,0.0); \fill[blue!50!black] (0.8811,1.0) circle (1.8pt); \fill[blue!50!black] (0.7088,0.8) circle (1.8pt); \fill[blue!50!black] (0.617,0.6) circle (1.8pt); \fill[blue!50!black] (0.5491,0.4) circle (1.8pt); \fill[blue!50!black] (0.4338,0.2) circle (1.8pt); \fill[blue!50!black] (0.1474,0.0) circle (1.8pt);\end{tikzpicture} & \begin{tikzpicture}[x=1.2cm,y=0.8cm,baseline={([yshift=-0.6ex]current bounding box.center)}] \draw[very thin, gray!15] (0,-0.1) -- (0,1.0); \draw[very thin, gray!15] (0.5,-0.1) -- (0.5,1.0); \draw[very thin, gray!15] (1.0,-0.1) -- (1.0,1.0); \draw[thick, red!75!black] (0.4552,1.0) -- (0.438,0.8) -- (0.576,0.6) -- (0.7431,0.4) -- (0.5508,0.2) -- (0.5434,0.0); \fill[blue!50!black] (0.4552,1.0) circle (1.8pt); \fill[blue!50!black] (0.438,0.8) circle (1.8pt); \fill[blue!50!black] (0.576,0.6) circle (1.8pt); \fill[blue!50!black] (0.7431,0.4) circle (1.8pt); \fill[blue!50!black] (0.5508,0.2) circle (1.8pt); \fill[blue!50!black] (0.5434,0.0) circle (1.8pt);\end{tikzpicture} \\
\midrule
Drift & \begin{tikzpicture}[x=1.2cm,y=0.8cm,baseline={([yshift=-0.6ex]current bounding box.center)}] \draw[very thin, gray!15] (0,-0.1) -- (0,1.0); \draw[very thin, gray!15] (0.5,-0.1) -- (0.5,1.0); \draw[very thin, gray!15] (1.0,-0.1) -- (1.0,1.0); \draw[thick, red!75!black] (0.0,0.8) -- (0.0,0.6) -- (0.0,0.4) -- (0.0,0.2) -- (0.0,0.0); \fill[blue!50!black] (0.0,0.8) circle (1.8pt); \fill[blue!50!black] (0.0,0.6) circle (1.8pt); \fill[blue!50!black] (0.0,0.4) circle (1.8pt); \fill[blue!50!black] (0.0,0.2) circle (1.8pt); \fill[blue!50!black] (0.0,0.0) circle (1.8pt);\end{tikzpicture} & \begin{tikzpicture}[x=1.2cm,y=0.8cm,baseline={([yshift=-0.6ex]current bounding box.center)}] \draw[very thin, gray!15] (0,-0.1) -- (0,1.0); \draw[very thin, gray!15] (0.5,-0.1) -- (0.5,1.0); \draw[very thin, gray!15] (1.0,-0.1) -- (1.0,1.0); \draw[thick, red!75!black] (0.85,1.0) -- (0.6953,0.8) -- (0.6118,0.6) -- (0.5411,0.4) -- (0.4356,0.2) -- (0.1223,0.0); \fill[blue!50!black] (0.85,1.0) circle (1.8pt); \fill[blue!50!black] (0.6953,0.8) circle (1.8pt); \fill[blue!50!black] (0.6118,0.6) circle (1.8pt); \fill[blue!50!black] (0.5411,0.4) circle (1.8pt); \fill[blue!50!black] (0.4356,0.2) circle (1.8pt); \fill[blue!50!black] (0.1223,0.0) circle (1.8pt);\end{tikzpicture} & \begin{tikzpicture}[x=1.2cm,y=0.8cm,baseline={([yshift=-0.6ex]current bounding box.center)}] \draw[very thin, gray!15] (0,-0.1) -- (0,1.0); \draw[very thin, gray!15] (0.5,-0.1) -- (0.5,1.0); \draw[very thin, gray!15] (1.0,-0.1) -- (1.0,1.0); \draw[thick, red!75!black] (0.8632,1.0) -- (0.7087,0.8) -- (0.6282,0.6) -- (0.559,0.4) -- (0.4398,0.2) -- (0.1533,0.0); \fill[blue!50!black] (0.8632,1.0) circle (1.8pt); \fill[blue!50!black] (0.7087,0.8) circle (1.8pt); \fill[blue!50!black] (0.6282,0.6) circle (1.8pt); \fill[blue!50!black] (0.559,0.4) circle (1.8pt); \fill[blue!50!black] (0.4398,0.2) circle (1.8pt); \fill[blue!50!black] (0.1533,0.0) circle (1.8pt);\end{tikzpicture} & \begin{tikzpicture}[x=1.2cm,y=0.8cm,baseline={([yshift=-0.6ex]current bounding box.center)}] \draw[very thin, gray!15] (0,-0.1) -- (0,1.0); \draw[very thin, gray!15] (0.5,-0.1) -- (0.5,1.0); \draw[very thin, gray!15] (1.0,-0.1) -- (1.0,1.0); \draw[thick, red!75!black] (0.4151,1.0) -- (0.4413,0.8) -- (0.6227,0.6) -- (0.7719,0.4) -- (0.5364,0.2) -- (0.5289,0.0); \fill[blue!50!black] (0.4151,1.0) circle (1.8pt); \fill[blue!50!black] (0.4413,0.8) circle (1.8pt); \fill[blue!50!black] (0.6227,0.6) circle (1.8pt); \fill[blue!50!black] (0.7719,0.4) circle (1.8pt); \fill[blue!50!black] (0.5364,0.2) circle (1.8pt); \fill[blue!50!black] (0.5289,0.0) circle (1.8pt);\end{tikzpicture} \\
\midrule
Rolling Mean & \begin{tikzpicture}[x=1.2cm,y=0.8cm,baseline={([yshift=-0.6ex]current bounding box.center)}] \draw[very thin, gray!15] (0,-0.1) -- (0,1.0); \draw[very thin, gray!15] (0.5,-0.1) -- (0.5,1.0); \draw[very thin, gray!15] (1.0,-0.1) -- (1.0,1.0); \draw[thick, red!75!black] (0.0,0.8) -- (0.0,0.6) -- (0.0,0.4) -- (0.0,0.2) -- (0.0,0.0); \fill[blue!50!black] (0.0,0.8) circle (1.8pt); \fill[blue!50!black] (0.0,0.6) circle (1.8pt); \fill[blue!50!black] (0.0,0.4) circle (1.8pt); \fill[blue!50!black] (0.0,0.2) circle (1.8pt); \fill[blue!50!black] (0.0,0.0) circle (1.8pt);\end{tikzpicture} & \begin{tikzpicture}[x=1.2cm,y=0.8cm,baseline={([yshift=-0.6ex]current bounding box.center)}] \draw[very thin, gray!15] (0,-0.1) -- (0,1.0); \draw[very thin, gray!15] (0.5,-0.1) -- (0.5,1.0); \draw[very thin, gray!15] (1.0,-0.1) -- (1.0,1.0); \draw[thick, red!75!black] (0.93,1.0) -- (0.7016,0.8) -- (0.5885,0.6) -- (0.5076,0.4) -- (0.4369,0.2) -- (0.1294,0.0); \fill[blue!50!black] (0.93,1.0) circle (1.8pt); \fill[blue!50!black] (0.7016,0.8) circle (1.8pt); \fill[blue!50!black] (0.5885,0.6) circle (1.8pt); \fill[blue!50!black] (0.5076,0.4) circle (1.8pt); \fill[blue!50!black] (0.4369,0.2) circle (1.8pt); \fill[blue!50!black] (0.1294,0.0) circle (1.8pt);\end{tikzpicture} & \begin{tikzpicture}[x=1.2cm,y=0.8cm,baseline={([yshift=-0.6ex]current bounding box.center)}] \draw[very thin, gray!15] (0,-0.1) -- (0,1.0); \draw[very thin, gray!15] (0.5,-0.1) -- (0.5,1.0); \draw[very thin, gray!15] (1.0,-0.1) -- (1.0,1.0); \draw[thick, red!75!black] (0.9361,1.0) -- (0.7117,0.8) -- (0.5938,0.6) -- (0.5146,0.4) -- (0.439,0.2) -- (0.1648,0.0); \fill[blue!50!black] (0.9361,1.0) circle (1.8pt); \fill[blue!50!black] (0.7117,0.8) circle (1.8pt); \fill[blue!50!black] (0.5938,0.6) circle (1.8pt); \fill[blue!50!black] (0.5146,0.4) circle (1.8pt); \fill[blue!50!black] (0.439,0.2) circle (1.8pt); \fill[blue!50!black] (0.1648,0.0) circle (1.8pt);\end{tikzpicture} & \begin{tikzpicture}[x=1.2cm,y=0.8cm,baseline={([yshift=-0.6ex]current bounding box.center)}] \draw[very thin, gray!15] (0,-0.1) -- (0,1.0); \draw[very thin, gray!15] (0.5,-0.1) -- (0.5,1.0); \draw[very thin, gray!15] (1.0,-0.1) -- (1.0,1.0); \draw[thick, red!75!black] (0.5615,1.0) -- (0.4547,0.8) -- (0.5235,0.6) -- (0.6262,0.4) -- (0.6295,0.2) -- (0.6274,0.0); \fill[blue!50!black] (0.5615,1.0) circle (1.8pt); \fill[blue!50!black] (0.4547,0.8) circle (1.8pt); \fill[blue!50!black] (0.5235,0.6) circle (1.8pt); \fill[blue!50!black] (0.6262,0.4) circle (1.8pt); \fill[blue!50!black] (0.6295,0.2) circle (1.8pt); \fill[blue!50!black] (0.6274,0.0) circle (1.8pt);\end{tikzpicture} \\
\midrule
Linear Regression & \begin{tikzpicture}[x=1.2cm,y=0.8cm,baseline={([yshift=-0.6ex]current bounding box.center)}] \draw[very thin, gray!15] (0,-0.1) -- (0,1.0); \draw[very thin, gray!15] (0.5,-0.1) -- (0.5,1.0); \draw[very thin, gray!15] (1.0,-0.1) -- (1.0,1.0); \draw[thick, red!75!black] (0.0,0.8) -- (0.2478,0.6) -- (0.2848,0.4) -- (0.3349,0.2) -- (0.3355,0.0); \fill[blue!50!black] (0.0,0.8) circle (1.8pt); \fill[blue!50!black] (0.2478,0.6) circle (1.8pt); \fill[blue!50!black] (0.2848,0.4) circle (1.8pt); \fill[blue!50!black] (0.3349,0.2) circle (1.8pt); \fill[blue!50!black] (0.3355,0.0) circle (1.8pt);\end{tikzpicture} & \begin{tikzpicture}[x=1.2cm,y=0.8cm,baseline={([yshift=-0.6ex]current bounding box.center)}] \draw[very thin, gray!15] (0,-0.1) -- (0,1.0); \draw[very thin, gray!15] (0.5,-0.1) -- (0.5,1.0); \draw[very thin, gray!15] (1.0,-0.1) -- (1.0,1.0); \draw[thick, red!75!black] (0.8622,1.0) -- (0.6889,0.8) -- (0.567,0.6) -- (0.4814,0.4) -- (0.4166,0.2) -- (0.1038,0.0); \fill[blue!50!black] (0.8622,1.0) circle (1.8pt); \fill[blue!50!black] (0.6889,0.8) circle (1.8pt); \fill[blue!50!black] (0.567,0.6) circle (1.8pt); \fill[blue!50!black] (0.4814,0.4) circle (1.8pt); \fill[blue!50!black] (0.4166,0.2) circle (1.8pt); \fill[blue!50!black] (0.1038,0.0) circle (1.8pt);\end{tikzpicture} & \begin{tikzpicture}[x=1.2cm,y=0.8cm,baseline={([yshift=-0.6ex]current bounding box.center)}] \draw[very thin, gray!15] (0,-0.1) -- (0,1.0); \draw[very thin, gray!15] (0.5,-0.1) -- (0.5,1.0); \draw[very thin, gray!15] (1.0,-0.1) -- (1.0,1.0); \draw[thick, red!75!black] (0.8743,1.0) -- (0.6966,0.8) -- (0.5588,0.6) -- (0.4723,0.4) -- (0.401,0.2) -- (0.1152,0.0); \fill[blue!50!black] (0.8743,1.0) circle (1.8pt); \fill[blue!50!black] (0.6966,0.8) circle (1.8pt); \fill[blue!50!black] (0.5588,0.6) circle (1.8pt); \fill[blue!50!black] (0.4723,0.4) circle (1.8pt); \fill[blue!50!black] (0.401,0.2) circle (1.8pt); \fill[blue!50!black] (0.1152,0.0) circle (1.8pt);\end{tikzpicture} & \begin{tikzpicture}[x=1.2cm,y=0.8cm,baseline={([yshift=-0.6ex]current bounding box.center)}] \draw[very thin, gray!15] (0,-0.1) -- (0,1.0); \draw[very thin, gray!15] (0.5,-0.1) -- (0.5,1.0); \draw[very thin, gray!15] (1.0,-0.1) -- (1.0,1.0); \draw[thick, red!75!black] (0.4341,1.0) -- (0.422,0.8) -- (0.4173,0.6) -- (0.4227,0.4) -- (0.4308,0.2) -- (0.4219,0.0); \fill[blue!50!black] (0.4341,1.0) circle (1.8pt); \fill[blue!50!black] (0.422,0.8) circle (1.8pt); \fill[blue!50!black] (0.4173,0.6) circle (1.8pt); \fill[blue!50!black] (0.4227,0.4) circle (1.8pt); \fill[blue!50!black] (0.4308,0.2) circle (1.8pt); \fill[blue!50!black] (0.4219,0.0) circle (1.8pt);\end{tikzpicture} \\
\midrule
ARIMAX & \begin{tikzpicture}[x=1.2cm,y=0.8cm,baseline={([yshift=-0.6ex]current bounding box.center)}] \draw[very thin, gray!15] (0,-0.1) -- (0,1.0); \draw[very thin, gray!15] (0.5,-0.1) -- (0.5,1.0); \draw[very thin, gray!15] (1.0,-0.1) -- (1.0,1.0); \draw[thick, red!75!black] (0.0,0.8) -- (0.1027,0.6) -- (0.0979,0.4) -- (0.2046,0.2) -- (0.2039,0.0); \fill[blue!50!black] (0.0,0.8) circle (1.8pt); \fill[blue!50!black] (0.1027,0.6) circle (1.8pt); \fill[blue!50!black] (0.0979,0.4) circle (1.8pt); \fill[blue!50!black] (0.2046,0.2) circle (1.8pt); \fill[blue!50!black] (0.2039,0.0) circle (1.8pt);\end{tikzpicture} & \begin{tikzpicture}[x=1.2cm,y=0.8cm,baseline={([yshift=-0.6ex]current bounding box.center)}] \draw[very thin, gray!15] (0,-0.1) -- (0,1.0); \draw[very thin, gray!15] (0.5,-0.1) -- (0.5,1.0); \draw[very thin, gray!15] (1.0,-0.1) -- (1.0,1.0); \draw[thick, red!75!black] (0.8117,1.0) -- (0.6932,0.8) -- (0.5805,0.6) -- (0.5021,0.4) -- (0.4324,0.2) -- (0.1199,0.0); \fill[blue!50!black] (0.8117,1.0) circle (1.8pt); \fill[blue!50!black] (0.6932,0.8) circle (1.8pt); \fill[blue!50!black] (0.5805,0.6) circle (1.8pt); \fill[blue!50!black] (0.5021,0.4) circle (1.8pt); \fill[blue!50!black] (0.4324,0.2) circle (1.8pt); \fill[blue!50!black] (0.1199,0.0) circle (1.8pt);\end{tikzpicture} & \begin{tikzpicture}[x=1.2cm,y=0.8cm,baseline={([yshift=-0.6ex]current bounding box.center)}] \draw[very thin, gray!15] (0,-0.1) -- (0,1.0); \draw[very thin, gray!15] (0.5,-0.1) -- (0.5,1.0); \draw[very thin, gray!15] (1.0,-0.1) -- (1.0,1.0); \draw[thick, red!75!black] (0.8283,1.0) -- (0.7032,0.8) -- (0.5765,0.6) -- (0.4912,0.4) -- (0.4165,0.2) -- (0.1313,0.0); \fill[blue!50!black] (0.8283,1.0) circle (1.8pt); \fill[blue!50!black] (0.7032,0.8) circle (1.8pt); \fill[blue!50!black] (0.5765,0.6) circle (1.8pt); \fill[blue!50!black] (0.4912,0.4) circle (1.8pt); \fill[blue!50!black] (0.4165,0.2) circle (1.8pt); \fill[blue!50!black] (0.1313,0.0) circle (1.8pt);\end{tikzpicture} & \begin{tikzpicture}[x=1.2cm,y=0.8cm,baseline={([yshift=-0.6ex]current bounding box.center)}] \draw[very thin, gray!15] (0,-0.1) -- (0,1.0); \draw[very thin, gray!15] (0.5,-0.1) -- (0.5,1.0); \draw[very thin, gray!15] (1.0,-0.1) -- (1.0,1.0); \draw[thick, red!75!black] (0.3626,1.0) -- (0.4577,0.8) -- (0.4969,0.6) -- (0.6144,0.4) -- (0.6411,0.2) -- (0.6429,0.0); \fill[blue!50!black] (0.3626,1.0) circle (1.8pt); \fill[blue!50!black] (0.4577,0.8) circle (1.8pt); \fill[blue!50!black] (0.4969,0.6) circle (1.8pt); \fill[blue!50!black] (0.6144,0.4) circle (1.8pt); \fill[blue!50!black] (0.6411,0.2) circle (1.8pt); \fill[blue!50!black] (0.6429,0.0) circle (1.8pt);\end{tikzpicture} \\
\midrule
SARIMAX & \begin{tikzpicture}[x=1.2cm,y=0.8cm,baseline={([yshift=-0.6ex]current bounding box.center)}] \draw[very thin, gray!15] (0,-0.1) -- (0,1.0); \draw[very thin, gray!15] (0.5,-0.1) -- (0.5,1.0); \draw[very thin, gray!15] (1.0,-0.1) -- (1.0,1.0); \draw[thick, red!75!black] (0.0,0.8) -- (0.1117,0.6) -- (0.0936,0.4) -- (0.1849,0.2) -- (0.1812,0.0); \fill[blue!50!black] (0.0,0.8) circle (1.8pt); \fill[blue!50!black] (0.1117,0.6) circle (1.8pt); \fill[blue!50!black] (0.0936,0.4) circle (1.8pt); \fill[blue!50!black] (0.1849,0.2) circle (1.8pt); \fill[blue!50!black] (0.1812,0.0) circle (1.8pt);\end{tikzpicture} & \begin{tikzpicture}[x=1.2cm,y=0.8cm,baseline={([yshift=-0.6ex]current bounding box.center)}] \draw[very thin, gray!15] (0,-0.1) -- (0,1.0); \draw[very thin, gray!15] (0.5,-0.1) -- (0.5,1.0); \draw[very thin, gray!15] (1.0,-0.1) -- (1.0,1.0); \draw[thick, red!75!black] (0.7042,0.8) -- (0.5725,0.6) -- (0.4991,0.4) -- (0.4259,0.2) -- (0.114,0.0); \fill[blue!50!black] (0.7042,0.8) circle (1.8pt); \fill[blue!50!black] (0.5725,0.6) circle (1.8pt); \fill[blue!50!black] (0.4991,0.4) circle (1.8pt); \fill[blue!50!black] (0.4259,0.2) circle (1.8pt); \fill[blue!50!black] (0.114,0.0) circle (1.8pt);\end{tikzpicture} & \begin{tikzpicture}[x=1.2cm,y=0.8cm,baseline={([yshift=-0.6ex]current bounding box.center)}] \draw[very thin, gray!15] (0,-0.1) -- (0,1.0); \draw[very thin, gray!15] (0.5,-0.1) -- (0.5,1.0); \draw[very thin, gray!15] (1.0,-0.1) -- (1.0,1.0); \draw[thick, red!75!black] (0.7114,0.8) -- (0.5698,0.6) -- (0.5036,0.4) -- (0.4178,0.2) -- (0.1312,0.0); \fill[blue!50!black] (0.7114,0.8) circle (1.8pt); \fill[blue!50!black] (0.5698,0.6) circle (1.8pt); \fill[blue!50!black] (0.5036,0.4) circle (1.8pt); \fill[blue!50!black] (0.4178,0.2) circle (1.8pt); \fill[blue!50!black] (0.1312,0.0) circle (1.8pt);\end{tikzpicture} & \begin{tikzpicture}[x=1.2cm,y=0.8cm,baseline={([yshift=-0.6ex]current bounding box.center)}] \draw[very thin, gray!15] (0,-0.1) -- (0,1.0); \draw[very thin, gray!15] (0.5,-0.1) -- (0.5,1.0); \draw[very thin, gray!15] (1.0,-0.1) -- (1.0,1.0); \draw[thick, red!75!black] (0.4695,0.8) -- (0.4736,0.6) -- (0.576,0.4) -- (0.5527,0.2) -- (0.5711,0.0); \fill[blue!50!black] (0.4695,0.8) circle (1.8pt); \fill[blue!50!black] (0.4736,0.6) circle (1.8pt); \fill[blue!50!black] (0.576,0.4) circle (1.8pt); \fill[blue!50!black] (0.5527,0.2) circle (1.8pt); \fill[blue!50!black] (0.5711,0.0) circle (1.8pt);\end{tikzpicture} \\
\midrule
Holt-Winters & \begin{tikzpicture}[x=1.2cm,y=0.8cm,baseline={([yshift=-0.6ex]current bounding box.center)}] \draw[very thin, gray!15] (0,-0.1) -- (0,1.0); \draw[very thin, gray!15] (0.5,-0.1) -- (0.5,1.0); \draw[very thin, gray!15] (1.0,-0.1) -- (1.0,1.0); \draw[thick, red!75!black] (0.0,0.8) -- (0.1018,0.6) -- (0.062,0.4) -- (0.0,0.2) -- (0.0,0.0); \fill[blue!50!black] (0.0,0.8) circle (1.8pt); \fill[blue!50!black] (0.1018,0.6) circle (1.8pt); \fill[blue!50!black] (0.062,0.4) circle (1.8pt); \fill[blue!50!black] (0.0,0.2) circle (1.8pt); \fill[red!80!black] (0.0,0.0) circle (1.8pt);\end{tikzpicture} & \begin{tikzpicture}[x=1.2cm,y=0.8cm,baseline={([yshift=-0.6ex]current bounding box.center)}] \draw[very thin, gray!15] (0,-0.1) -- (0,1.0); \draw[very thin, gray!15] (0.5,-0.1) -- (0.5,1.0); \draw[very thin, gray!15] (1.0,-0.1) -- (1.0,1.0); \draw[thick, red!75!black] (0.9515,1.0) -- (0.7281,0.8) -- (0.5813,0.6) -- (0.5092,0.4) -- (0.5213,0.2) -- (0.443,0.0); \fill[blue!50!black] (0.9515,1.0) circle (1.8pt); \fill[blue!50!black] (0.7281,0.8) circle (1.8pt); \fill[blue!50!black] (0.5813,0.6) circle (1.8pt); \fill[blue!50!black] (0.5092,0.4) circle (1.8pt); \fill[blue!50!black] (0.5213,0.2) circle (1.8pt); \fill[blue!50!black] (0.443,0.0) circle (1.8pt);\end{tikzpicture} & \begin{tikzpicture}[x=1.2cm,y=0.8cm,baseline={([yshift=-0.6ex]current bounding box.center)}] \draw[very thin, gray!15] (0,-0.1) -- (0,1.0); \draw[very thin, gray!15] (0.5,-0.1) -- (0.5,1.0); \draw[very thin, gray!15] (1.0,-0.1) -- (1.0,1.0); \draw[thick, red!75!black] (0.9558,1.0) -- (0.7316,0.8) -- (0.5836,0.6) -- (0.5034,0.4) -- (0.5289,0.2) -- (0.4736,0.0); \fill[blue!50!black] (0.9558,1.0) circle (1.8pt); \fill[blue!50!black] (0.7316,0.8) circle (1.8pt); \fill[blue!50!black] (0.5836,0.6) circle (1.8pt); \fill[blue!50!black] (0.5034,0.4) circle (1.8pt); \fill[blue!50!black] (0.5289,0.2) circle (1.8pt); \fill[blue!50!black] (0.4736,0.0) circle (1.8pt);\end{tikzpicture} & \begin{tikzpicture}[x=1.2cm,y=0.8cm,baseline={([yshift=-0.6ex]current bounding box.center)}] \draw[very thin, gray!15] (0,-0.1) -- (0,1.0); \draw[very thin, gray!15] (0.5,-0.1) -- (0.5,1.0); \draw[very thin, gray!15] (1.0,-0.1) -- (1.0,1.0); \draw[thick, red!75!black] (0.69,1.0) -- (0.4995,0.8) -- (0.4705,0.6) -- (0.5255,0.4) -- (0.8648,0.2) -- (1.0,0.0); \fill[blue!50!black] (0.69,1.0) circle (1.8pt); \fill[blue!50!black] (0.4995,0.8) circle (1.8pt); \fill[blue!50!black] (0.4705,0.6) circle (1.8pt); \fill[blue!50!black] (0.5255,0.4) circle (1.8pt); \fill[blue!50!black] (0.8648,0.2) circle (1.8pt); \fill[red!80!black] (1.0,0.0) circle (1.8pt);\end{tikzpicture} \\
\midrule
XGBoost & \begin{tikzpicture}[x=1.2cm,y=0.8cm,baseline={([yshift=-0.6ex]current bounding box.center)}] \draw[very thin, gray!15] (0,-0.1) -- (0,1.0); \draw[very thin, gray!15] (0.5,-0.1) -- (0.5,1.0); \draw[very thin, gray!15] (1.0,-0.1) -- (1.0,1.0); \draw[thick, red!75!black] (0.0,0.8) -- (0.0,0.6) -- (0.0904,0.4) -- (0.1429,0.2) -- (0.5902,0.0); \fill[blue!50!black] (0.0,0.8) circle (1.8pt); \fill[blue!50!black] (0.0,0.6) circle (1.8pt); \fill[blue!50!black] (0.0904,0.4) circle (1.8pt); \fill[blue!50!black] (0.1429,0.2) circle (1.8pt); \fill[blue!50!black] (0.5902,0.0) circle (1.8pt);\end{tikzpicture} & \begin{tikzpicture}[x=1.2cm,y=0.8cm,baseline={([yshift=-0.6ex]current bounding box.center)}] \draw[very thin, gray!15] (0,-0.1) -- (0,1.0); \draw[very thin, gray!15] (0.5,-0.1) -- (0.5,1.0); \draw[very thin, gray!15] (1.0,-0.1) -- (1.0,1.0); \draw[thick, red!75!black] (0.9269,1.0) -- (0.7446,0.8) -- (0.5995,0.6) -- (0.5021,0.4) -- (0.4433,0.2) -- (0.0601,0.0); \fill[blue!50!black] (0.9269,1.0) circle (1.8pt); \fill[blue!50!black] (0.7446,0.8) circle (1.8pt); \fill[blue!50!black] (0.5995,0.6) circle (1.8pt); \fill[blue!50!black] (0.5021,0.4) circle (1.8pt); \fill[blue!50!black] (0.4433,0.2) circle (1.8pt); \fill[blue!50!black] (0.0601,0.0) circle (1.8pt);\end{tikzpicture} & \begin{tikzpicture}[x=1.2cm,y=0.8cm,baseline={([yshift=-0.6ex]current bounding box.center)}] \draw[very thin, gray!15] (0,-0.1) -- (0,1.0); \draw[very thin, gray!15] (0.5,-0.1) -- (0.5,1.0); \draw[very thin, gray!15] (1.0,-0.1) -- (1.0,1.0); \draw[thick, red!75!black] (0.9333,1.0) -- (0.749,0.8) -- (0.5998,0.6) -- (0.4945,0.4) -- (0.4346,0.2) -- (0.0552,0.0); \fill[blue!50!black] (0.9333,1.0) circle (1.8pt); \fill[blue!50!black] (0.749,0.8) circle (1.8pt); \fill[blue!50!black] (0.5998,0.6) circle (1.8pt); \fill[blue!50!black] (0.4945,0.4) circle (1.8pt); \fill[blue!50!black] (0.4346,0.2) circle (1.8pt); \fill[blue!50!black] (0.0552,0.0) circle (1.8pt);\end{tikzpicture} & \begin{tikzpicture}[x=1.2cm,y=0.8cm,baseline={([yshift=-0.6ex]current bounding box.center)}] \draw[very thin, gray!15] (0,-0.1) -- (0,1.0); \draw[very thin, gray!15] (0.5,-0.1) -- (0.5,1.0); \draw[very thin, gray!15] (1.0,-0.1) -- (1.0,1.0); \draw[thick, red!75!black] (0.581,1.0) -- (0.6057,0.8) -- (0.5016,0.6) -- (0.5617,0.4) -- (0.5683,0.2) -- (0.4377,0.0); \fill[blue!50!black] (0.581,1.0) circle (1.8pt); \fill[blue!50!black] (0.6057,0.8) circle (1.8pt); \fill[blue!50!black] (0.5016,0.6) circle (1.8pt); \fill[blue!50!black] (0.5617,0.4) circle (1.8pt); \fill[blue!50!black] (0.5683,0.2) circle (1.8pt); \fill[blue!50!black] (0.4377,0.0) circle (1.8pt);\end{tikzpicture} \\
\midrule
LSTM PyTorch & \begin{tikzpicture}[x=1.2cm,y=0.8cm,baseline={([yshift=-0.6ex]current bounding box.center)}] \draw[very thin, gray!15] (0,-0.1) -- (0,1.0); \draw[very thin, gray!15] (0.5,-0.1) -- (0.5,1.0); \draw[very thin, gray!15] (1.0,-0.1) -- (1.0,1.0); \draw[thick, red!75!black] (0.0,0.8) -- (0.0663,0.6) -- (0.019,0.4) -- (0.1813,0.2) -- (1.0,0.0); \fill[blue!50!black] (0.0,0.8) circle (1.8pt); \fill[blue!50!black] (0.0663,0.6) circle (1.8pt); \fill[blue!50!black] (0.019,0.4) circle (1.8pt); \fill[blue!50!black] (0.1813,0.2) circle (1.8pt); \fill[green!60!black] (1.0,0.0) circle (1.8pt);\end{tikzpicture} & \begin{tikzpicture}[x=1.2cm,y=0.8cm,baseline={([yshift=-0.6ex]current bounding box.center)}] \draw[very thin, gray!15] (0,-0.1) -- (0,1.0); \draw[very thin, gray!15] (0.5,-0.1) -- (0.5,1.0); \draw[very thin, gray!15] (1.0,-0.1) -- (1.0,1.0); \draw[thick, red!75!black] (1.0,1.0) -- (0.7398,0.8) -- (0.5824,0.6) -- (0.5179,0.4) -- (0.4384,0.2) -- (0.0,0.0); \fill[red!80!black] (1.0,1.0) circle (1.8pt); \fill[blue!50!black] (0.7398,0.8) circle (1.8pt); \fill[blue!50!black] (0.5824,0.6) circle (1.8pt); \fill[blue!50!black] (0.5179,0.4) circle (1.8pt); \fill[blue!50!black] (0.4384,0.2) circle (1.8pt); \fill[green!60!black] (0.0,0.0) circle (1.8pt);\end{tikzpicture} & \begin{tikzpicture}[x=1.2cm,y=0.8cm,baseline={([yshift=-0.6ex]current bounding box.center)}] \draw[very thin, gray!15] (0,-0.1) -- (0,1.0); \draw[very thin, gray!15] (0.5,-0.1) -- (0.5,1.0); \draw[very thin, gray!15] (1.0,-0.1) -- (1.0,1.0); \draw[thick, red!75!black] (1.0,1.0) -- (0.7446,0.8) -- (0.5792,0.6) -- (0.5132,0.4) -- (0.4241,0.2) -- (0.0,0.0); \fill[red!80!black] (1.0,1.0) circle (1.8pt); \fill[blue!50!black] (0.7446,0.8) circle (1.8pt); \fill[blue!50!black] (0.5792,0.6) circle (1.8pt); \fill[blue!50!black] (0.5132,0.4) circle (1.8pt); \fill[blue!50!black] (0.4241,0.2) circle (1.8pt); \fill[green!60!black] (0.0,0.0) circle (1.8pt);\end{tikzpicture} & \begin{tikzpicture}[x=1.2cm,y=0.8cm,baseline={([yshift=-0.6ex]current bounding box.center)}] \draw[very thin, gray!15] (0,-0.1) -- (0,1.0); \draw[very thin, gray!15] (0.5,-0.1) -- (0.5,1.0); \draw[very thin, gray!15] (1.0,-0.1) -- (1.0,1.0); \draw[thick, red!75!black] (0.7195,1.0) -- (0.5839,0.8) -- (0.4813,0.6) -- (0.6736,0.4) -- (0.6615,0.2) -- (0.0,0.0); \fill[blue!50!black] (0.7195,1.0) circle (1.8pt); \fill[blue!50!black] (0.5839,0.8) circle (1.8pt); \fill[blue!50!black] (0.4813,0.6) circle (1.8pt); \fill[blue!50!black] (0.6736,0.4) circle (1.8pt); \fill[blue!50!black] (0.6615,0.2) circle (1.8pt); \fill[green!60!black] (0.0,0.0) circle (1.8pt);\end{tikzpicture} \\
\midrule
N-BEATS & \begin{tikzpicture}[x=1.2cm,y=0.8cm,baseline={([yshift=-0.6ex]current bounding box.center)}] \draw[very thin, gray!15] (0,-0.1) -- (0,1.0); \draw[very thin, gray!15] (0.5,-0.1) -- (0.5,1.0); \draw[very thin, gray!15] (1.0,-0.1) -- (1.0,1.0); \draw[thick, red!75!black] (0.0,0.8) -- (0.0931,0.6) -- (0.169,0.4) -- (0.2196,0.2) -- (0.2435,0.0); \fill[blue!50!black] (0.0,0.8) circle (1.8pt); \fill[blue!50!black] (0.0931,0.6) circle (1.8pt); \fill[blue!50!black] (0.169,0.4) circle (1.8pt); \fill[blue!50!black] (0.2196,0.2) circle (1.8pt); \fill[blue!50!black] (0.2435,0.0) circle (1.8pt);\end{tikzpicture} & \begin{tikzpicture}[x=1.2cm,y=0.8cm,baseline={([yshift=-0.6ex]current bounding box.center)}] \draw[very thin, gray!15] (0,-0.1) -- (0,1.0); \draw[very thin, gray!15] (0.5,-0.1) -- (0.5,1.0); \draw[very thin, gray!15] (1.0,-0.1) -- (1.0,1.0); \draw[thick, red!75!black] (0.7616,0.8) -- (0.5737,0.6) -- (0.4931,0.4) -- (0.4334,0.2) -- (0.1144,0.0); \fill[blue!50!black] (0.7616,0.8) circle (1.8pt); \fill[blue!50!black] (0.5737,0.6) circle (1.8pt); \fill[blue!50!black] (0.4931,0.4) circle (1.8pt); \fill[blue!50!black] (0.4334,0.2) circle (1.8pt); \fill[blue!50!black] (0.1144,0.0) circle (1.8pt);\end{tikzpicture} & \begin{tikzpicture}[x=1.2cm,y=0.8cm,baseline={([yshift=-0.6ex]current bounding box.center)}] \draw[very thin, gray!15] (0,-0.1) -- (0,1.0); \draw[very thin, gray!15] (0.5,-0.1) -- (0.5,1.0); \draw[very thin, gray!15] (1.0,-0.1) -- (1.0,1.0); \draw[thick, red!75!black] (0.7543,0.8) -- (0.5749,0.6) -- (0.5011,0.4) -- (0.4354,0.2) -- (0.1394,0.0); \fill[blue!50!black] (0.7543,0.8) circle (1.8pt); \fill[blue!50!black] (0.5749,0.6) circle (1.8pt); \fill[blue!50!black] (0.5011,0.4) circle (1.8pt); \fill[blue!50!black] (0.4354,0.2) circle (1.8pt); \fill[blue!50!black] (0.1394,0.0) circle (1.8pt);\end{tikzpicture} & \begin{tikzpicture}[x=1.2cm,y=0.8cm,baseline={([yshift=-0.6ex]current bounding box.center)}] \draw[very thin, gray!15] (0,-0.1) -- (0,1.0); \draw[very thin, gray!15] (0.5,-0.1) -- (0.5,1.0); \draw[very thin, gray!15] (1.0,-0.1) -- (1.0,1.0); \draw[thick, red!75!black] (0.4835,0.8) -- (0.4156,0.6) -- (0.3046,0.4) -- (0.3284,0.2) -- (0.2894,0.0); \fill[blue!50!black] (0.4835,0.8) circle (1.8pt); \fill[blue!50!black] (0.4156,0.6) circle (1.8pt); \fill[blue!50!black] (0.3046,0.4) circle (1.8pt); \fill[blue!50!black] (0.3284,0.2) circle (1.8pt); \fill[blue!50!black] (0.2894,0.0) circle (1.8pt);\end{tikzpicture} \\
\bottomrule
\end{tabular}%
}
\end{table}

\section{Discussion}

Several key statistical insights emerge from the empirical benchmarking:
\begin{enumerate}
\item \textbf{The In-Sample Fit Illusion}: Finer temporal resolution (Daily) yields high training $R^2$ values that bear no relationship to out-of-sample accuracy. A clear example is the naïve Persistence baseline: at the Daily grain it attains Train $R^2 = 0.7841$ --- comparable to the deep models --- yet its Test $R^2$ falls to $-1.12$ and its TPFE to $23.89\%$. The model captured the high-frequency autocorrelation while acquiring no out-of-sample forecasting skill. The effect is amplified for parameterized seasonal models: Holt-Winters at the Daily grain produces a Test $R^2$ of $-151$ and a TPFE of $425.85\%$, the worst outcome in the benchmark, because its seasonal state is recursively re-fed over $H=365$ steps. This constitutes evidence that high in-sample fit at fine grains is a structural predictor of out-of-sample degradation, not merely a heuristic warning.
\item \textbf{Autoregressive vs. Trend Stability}: Linear Regression remains stable across the granularity paradox, maintaining TPFE in the narrow band $16.09$--$16.96\%$ across all \emph{six} granularities (Annual through Daily, a $0.87$~pp spread). This occurs because the model projects predictions directly as a function of time, without the recursive feedback loop characteristic of autoregressive and state-updating models. The six-grain flatness --- confirmed across two intermediate granularities (Quarterly, Bi-Weekly) that were absent from prior work --- constitutes empirical evidence that the paradox is driven by recursive feedback topology, not by model complexity or sample size.
\item \textbf{The Non-Monotonic LSTM Threshold (U-Shaped Error Curve)}: The extended six-grain benchmark reveals that the LSTM's ``data hunger threshold'' is not a monotonic boundary but a U-shaped error curve. The LSTM's TPFE worsens from Monthly ($19.66\%$) through Quarterly ($27.12\%$) to Bi-Weekly ($35.94\%$), then partially recovers at Weekly ($34.59\%$) before decreasing to $4.35\%$ at Daily. The intermediate granularities form an adverse operating region: $N$ is larger than Monthly, but the recursive horizon $H$ grows faster than the data can compensate, so error propagation dominates. Only at Daily ($N=4754$) does the sample size finally resolve the LSTM's high-dimensional parameter space, yielding a Test $R^2$ of $0.66$. This non-monotonicity means that practitioners who choose an intermediate grain (Bi-Weekly, Weekly) thinking it ``balances'' data and horizon are in fact selecting the \emph{worst} operating point for recursive deep models.
\item \textbf{N-BEATS Robustness vs.~LSTM Sensitivity}: N-BEATS shows a flatter TPFE profile ($10.77$--$19.80\%$ across all fine grains) and already achieves $10.77\%$ at Daily and $11.30\%$ at Bi-Weekly --- competitive with the LSTM's Daily result, but without the intermediate-grain degradation. Its trend-block inductive bias appears to resolve the multi-step dependency without requiring the full Daily sample, whereas the LSTM's recurrent hidden state accumulates error linearly with $H$ until the data volume is large enough to compensate.
\item \textbf{Metric Consensus vs. Cumulative Dissensus}: Because the objective of temporal aggregation in this context is cumulative planning and budgeting, TPFE serves as the step-agnostic, goal-dependent primary evaluation criterion. Analyzing the consensus between pointwise metrics (RMSE, MAE, $R^2$) and TPFE reveals the nature of a model's error propagation. For non-recursive models (Linear Regression), there is perfect consensus: both pointwise and cumulative metrics agree on stability. In contrast, recursive autoregressive models (e.g., SARIMAX, ARIMAX) exhibit a clear dissensus: pointwise metrics (RMSE, MAE) improve or stabilize at finer grains (due to sample size expansion reducing parameter variance), yet cumulative TPFE degrades. This dissensus exposes systematic forecasting bias that propagates recursively over the daily planning horizon ($H=365$), which pointwise metrics hide. Conversely, unbiased high-variance models benefit from error cancellation over long planning periods, where positive and negative daily fluctuations cancel out, yielding a low TPFE despite high high-frequency pointwise variance (as seen in the daily LSTM).
\end{enumerate}

\FloatBarrier

\begin{table}[!htbp]
\centering
\caption{Consensus and Dissensus Classifications between Pointwise Fit and Cumulative Budgeting Error (Daily vs. Monthly resolution)}
\label{tab:consensus_summary}
\small
\resizebox{\columnwidth}{!}{%
\begin{tabular}{lllll}
\toprule
\textbf{Model Class} & \textbf{Model} & \textbf{Pointwise Trend} & \textbf{TPFE Trend} & \textbf{Classification} \\
\midrule
Naïve (Recursive) & Persistence & In-Sample Fit Illusion & Degraded & \textbf{Dissensus} \\
Naïve (Drift) & Drift & Poor / Flat & Moderate & \textbf{Dissensus} \\
Naïve (Non-recursive) & Rolling Mean & Moderate & Degraded & \textbf{Dissensus} \\
Statistical (Linear) & Linear Regression & Stable / Scale-invariant & Stable & \textbf{Consensus} \\
Autoregressive & ARIMAX & Pointwise Improvement & Degraded & \textbf{Dissensus} \\
Autoregressive & SARIMAX & Pointwise Improvement & Degraded & \textbf{Dissensus} \\
Exponential Smoothing & Holt-Winters & Catastrophic Collapse & Catastrophic & \textbf{Dissensus} \\
Machine Learning & XGBoost & Clear Improvement & Improved & \textbf{Consensus} \\
Deep Learning & LSTM PyTorch & Clear Improvement & Clear Improvement & \textbf{Consensus} \\
Deep Learning & N-BEATS & Stable/Improvement & Stable/Improved & \textbf{Consensus} \\
\bottomrule
\end{tabular}%
}
\end{table}

\FloatBarrier

\section{Conclusion}
The choice of temporal grain is not merely a matter of data representation, but a design decision that controls error dynamics. We summarize three methodological conclusions that emerge from the empirical results.

\textbf{Pointwise metrics are insufficient for cross-granularity evaluation.} RMSE and MAE decrease monotonically as temporal resolution increases---Holt-Winters Daily records a Test RMSE of 5.34M versus 12.62M at Monthly---yet its cumulative TPFE rises from $19.01\%$ to $425.85\%$. This occurs because RMSE and MAE evaluate individual step accuracy within the observation window of each granularity, not the accumulated deviation over the planning horizon. These metrics remain useful for diagnosing point-by-point prediction quality at a fixed granularity, but they are uninformative---and potentially misleading---when used to compare model adequacy across granularities or to assess cumulative planning accuracy.

\textbf{Forecasting without a goal-dependent cumulative metric creates distorted evaluation.} Across the benchmark, models evaluated exclusively on in-sample $R^2$, RMSE, and MAE would rank Persistence at Daily as a competitive forecaster (Train $R^2 = 0.78$, low RMSE). Only when evaluated against TPFE---a metric anchored to the actual planning objective---does the model's inadequacy become visible. This finding generalizes beyond our specific dataset: any forecasting exercise that lacks a metric tied to the downstream decision variable (cumulative budget deviation, total demand, aggregate volume) risks producing models that perform well on diagnostic criteria while failing the task they were built for.

\textbf{In-sample-only evaluation creates a perverse incentive structure.} In-sample fit diagnostics systematically reward two choices: finer temporal disaggregation (which inflates $N$ and improves training statistics) and greater model complexity (which reduces training residuals). Both choices increase the recursive forecasting horizon $H$ or the parameter space, amplifying out-of-sample error propagation. A workflow that monitors only in-sample metrics therefore incentivizes precisely the design decisions that degrade forecasting performance. This is not a theoretical concern: the benchmark shows that the model--granularity combination with the highest training $R^2$ (Persistence at Daily, $0.78$) is outperformed out-of-sample by Linear Regression ($16.32\%$ TPFE vs.~$23.89\%$), a model whose training $R^2$ ($0.15$) would appear inferior by in-sample standards.

From a practical standpoint, finer granularities should be selected for high-capacity models (such as LSTMs) only when the resulting sample size crosses the threshold required to resolve the architecture's parameter space. For low-data regimes, non-recursive trend models or parsimonious monthly statistical models remain preferable. In all cases, evaluation should include at least one goal-dependent cumulative metric aligned with the planning objective.

\section{Limitations and Reproducibility}
The benchmark is restricted to a single series (CPV Division 72, IT Services), so the absolute TPFE magnitudes are dataset-specific. The directional findings --- recursive degradation of autoregressive models at fine grains, U-shaped LSTM error curve, Linear Regression scale-invariance --- are structural properties of the model--granularity interaction and should generalise, but confirming this across additional CPV divisions is left for future work. Confidence intervals from multi-seed runs and Diebold--Mariano significance tests between models are also deferred: the present results are single-seed point estimates and should be read as a structural comparison rather than a ranking.

The dataset is derived from the Portuguese public contracts registry (Portal BASE), which is publicly accessible at \url{https://www.base.gov.pt}. The CPV Division 72 filter and the resampling protocol are described in Section~\ref{sec:empirical} (Empirical Evaluation); full pre-processing code, model configurations, and the backtesting pipeline are available from the author on request. Software stack: Python 3.11, PyTorch 2.x, Darts, statsmodels, scikit-learn, XGBoost.

\FloatBarrier
\clearpage

\bibliographystyle{plainnat}
\bibliography{references}

\end{document}